\documentclass[12pt,onecolumn,letterpaper]{article}

\usepackage{cvpr}
\usepackage{times}
\usepackage{epsfig}
\usepackage{graphicx}
\usepackage{amsmath}
\usepackage{amssymb}
\usepackage{multirow}
\usepackage{algorithm}
\usepackage{algorithmic}
\usepackage{multirow}
\usepackage{wrapfig}
\usepackage{subfig}
\usepackage{array}

\newcolumntype{C}[1]{>{\centering\let\newline\\\arraybackslash\hspace{0pt}}m{#1}}
\usepackage{hyperref}

 \cvprfinalcopy 

\newcommand{\norm}[1]{\left\lVert#1\right\rVert} 
\def\Vec#1{{\boldsymbol{#1}}}
\def\Mat#1{{\boldsymbol{#1}}}

\DeclareMathOperator*{\argmin}{arg\,min}

\ifcvprfinal\fi
\begin{document}

\title{Using LIP to Gloss Over Faces in Single-Stage Face Detection Networks}

\author{Siqi Yang, Arnold Wiliem, Shaokang Chen, Brian C. Lovell \\
The University of Queensland, School of ITEE, QLD 4072, Australia\\
{\tt\small \{siqi.yang, a.wiliem, s.chen2\}@uq.edu.au, lovell@itee.uq.edu.au}
}

\maketitle

\begin{abstract}
This work shows that it is possible to fool/attack
recent state-of-the-art face detectors which are 
based on the single-stage networks.
Successfully attacking face detectors could be a 
serious malware vulnerability when deploying a smart surveillance system 
utilizing face detectors.
In addition, for the privacy concern, it helps prevent faces being harvested and stored in the server.
We show that existing adversarial perturbation methods
are not effective to perform such an attack, especially
when there are multiple faces in the input image.
This is because the adversarial perturbation specifically generated
for one face may disrupt the adversarial perturbation for another face.
In this paper, we call this problem the Instance Perturbation
Interference (IPI) problem.
This IPI problem is addressed by studying the relationship 
between the deep neural network receptive field and 
the adversarial perturbation.
As such, we propose the Localized Instance Perturbation (LIP)
that confines the adversarial perturbation inside the Effective Receptive
Field (ERF) of a target to perform the attack.
Experimental results show the LIP method massively outperforms existing adversarial 
perturbation generation methods -- often by a factor of 2 to 10.

\end{abstract}


\section{Introduction}

Deep neural networks have achieved great success in recent years on many
applications~\cite{VGG,AlexNet,ResNet,fasterRCNN,SSD,YOLO,Tinyfaces,MTCNN,FCN}. However, it has been demonstrated in various
works that by adding tiny, imperceptible perturbations onto the image,
the network output can be changed significantly~\cite{intriguing,FGSM,delving,easilyfooled,deepfool,adversariaPhysical,DAG,universaSegmentation}.  These
perturbations are often referred to as adversarial
perturbations~\cite{FGSM}.
Most prior works are primarily aimed at generating adversarial perturbations to fool neural networks for image classification tasks~\cite{intriguing,FGSM,delving,easilyfooled,deepfool,adversariaPhysical,universal}.
It is relatively easier to attack these networks as the perturbations need to change only one network decision for each image containing an instance/object of interest. This means, there is only a single target and the target is the entire image.
Recently, several methods have been proposed on more challenging attacks for segmentation~\cite{universaSegmentation,adversarialSSICLRwl,houdini} and object detection tasks~\cite{DAG}, where there are significantly more targets to attack within the input image. 



In the field of biometrics, Sharif~\etal\cite{adversarialGlasses} showed that face recognition systems can be fooled by applying adversarial perturbations, where a detected face can be recognized as another individual.
In addition, for the privacy concern, biometric data in a dataset might be utilized without the consent of the users.
Therefore, Mirjalili~\etal\cite{SemiAdverssarialNet,SoftBiometricPrivacy} developed a technique to protect the soft biometric privacy (\eg gender) without harming the accuracy of face recognition.
However, in the above-mentioned methods, the faces are still captured and stored in a server.
In this paper, we propose a novel way to address these privacy issues by avoiding the faces be detected completely from an image.
Thus, attacking face detection is crucial for both the security and privacy concerns.

With similar goals, previous works~\cite{adversarialGlasses,faceAttackLight} performed attacks on the Viola and Jones (VJ) face detector~\cite{VJ}. 
However, deep neural networks have been shown to be extremely effective in detecting faces~\cite{Tinyfaces,MTCNN,CnnCascade,JointCNN,Faceness,STN,Conv3D,SSH,S3FD,JDA}, which can achieve 2 times higher detection rate than the VJ. 
In this work, we tackle the problem of generating effective adversarial perturbations for deep learning based face detection networks. 
To the best of our knowledge, this is the first study that attempts to perform such an adversarial attack on face detection networks.

In recent years, deep network based object/face detection methods can be grouped as two-stage network, \eg Faster-RCNN~\cite{fasterRCNN} and single-stage network~\cite{SSD,YOLO,Tinyfaces,SSH,S3FD}. In Faster-RCNN~\cite{faceFasterRCNN}, a shallow region proposal network is applied
to generate candidates and a deep classification network is utilized for the final decision. 
The Single-Stage (SS) network is similar to the region proposal network in Faster-RCNN~\cite{fasterRCNN} but performs both object classification and localization simultaneously.
By utilizing the Single-Stage network architecture, recent detectors~\cite{Tinyfaces,SSH,S3FD} can detect faces on various scales
with a much faster running time. 
Due to their excellent performance, we confine
this paper to attacking the most recent face detectors utilizing
Single-Stage network.

\begin{figure}[t]
\begin{center}
   \includegraphics[width=0.8\linewidth]{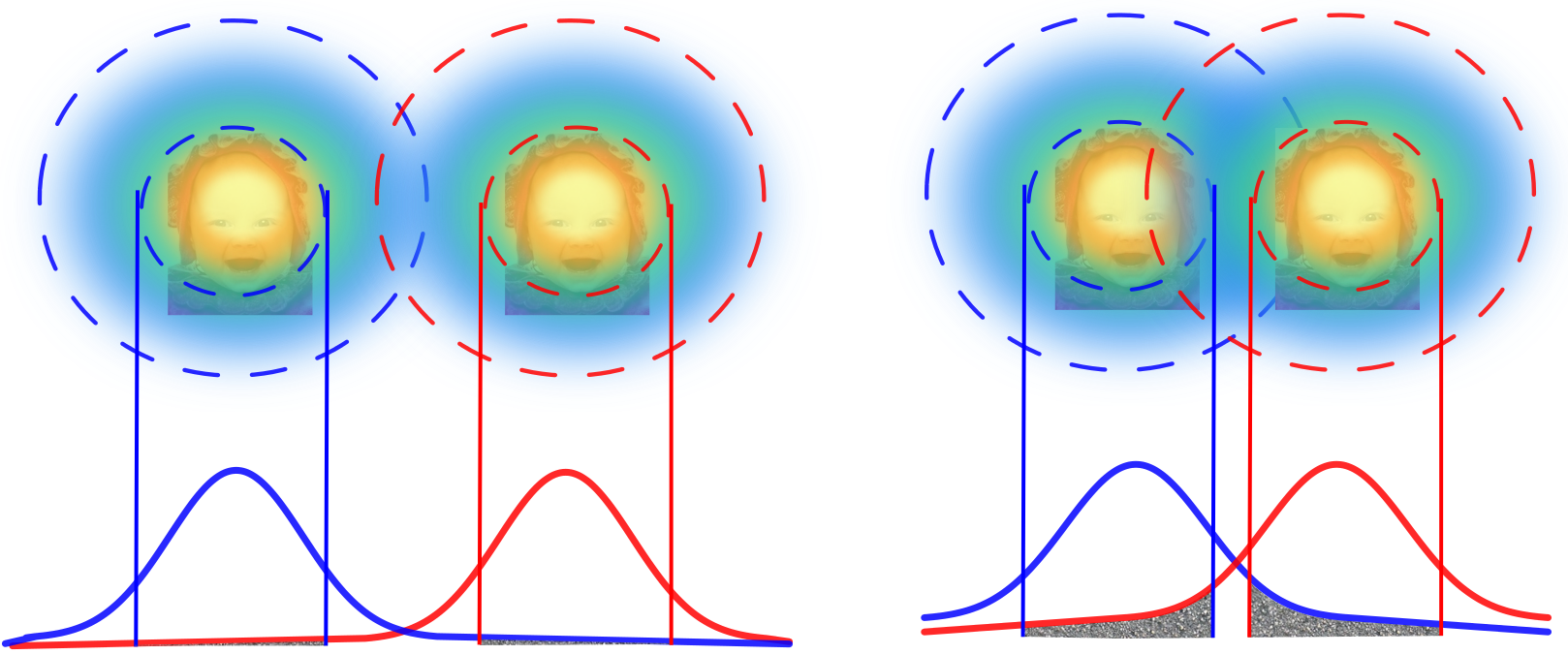}
\end{center}
\vspace{-10pt}
   \caption{An illustration of the Instance Perturbation Interference (IPI) problem. 
   Upper row: two instances with their generated adversarial perturbations.
   The outer and inner circles indicate the Theoretical Receptive Field (TRF) and Effective Receptive Field (ERF), respectively.
   Lower row: one dimensional representation of the perturbations. IPI problem refers to the perturbation generated for one instance
   significantly disrupting the perturbation generated for the other instance. 
   The disruption does not have significant effect on the left case, whereas on the right case, it will reduce the effectiveness of the attack.}
\label{fig:neighbour-interference}
\vspace{-10pt}
\end{figure}

We find that applying the commonly-used gradient based adversarial methods~\cite{FGSM,deepfool} to the state-of-the-art face detection networks has not presented satisfactory results.
We point out that attacking a Single-Stage detector is challenging and the unsatisfactory performance is attributed to the Instance Perturbation Interference
(IPI) problem. 
The IPI problem can be briefly explained as interference between the perturbation required to attack one instance and the perturbation required to attack a nearby instance. 
Since the recent adversarial perturbation methods~\cite{DAG,universaSegmentation} do not
consider this problem, they become quite ineffective in attacking SS face detector networks.




In this work, we attribute the IPI problem to the receptive field of deep
neural networks. Recent work~\cite{ERF} shows that the receptive field follows a 
2D Gaussian distribution, where the set of input image pixels closer to an output neuron have higher impact on the neuron decision. The
area where high impact pixels are concentrated is referred to as the
Effective Receptive Field (ERF)~\cite{ERF}. As illustrated in
Fig.~\ref{fig:neighbour-interference}, if two faces are close to
each other, the perturbation generated to attack one face will reside
in the ERF of another face.
Prior work~\cite{mitigatingRandomization} shows that adversarial attacks might fail when the specific structure is destroyed.
Thus, the residency in the ERF significantly hampers the success of attacking the other face.
In other words, the IPI problem happens when the interfering perturbations
disrupt the adversarial perturbations generated for the neighboring
faces. This IPI problem will become more serious when multiple faces
exist in close proximity. We believe this
is the first work that describes and explains the IPI problem.
\noindent \textbf{Contributions - } We list our contributions as
follows: (1) We describe and provide theoretical explanation of the Instance Perturbation Interference
problem that makes the existing adversarial perturbation
generation method fail to attack the SS face detector networks when
multiple faces exist;
(2) This is the first study to show that it is
possible to attack deep neural network based face detector. More
specifically, we propose an approach to attack Single-Stage
based face detector networks. (3) To perform the attack, we propose Localised Instance Perturbation (LIP) method to generate instance based perturbations by confining the perturbations inside each instance ERF. 


We continue our paper as follows. We discuss related work and background in Section 2. Then we describe the Instance Perturbation Interference problem in Section 3. The proposed Localized Instance Perturbation is described in Section 4 followed by the validation experiments in Section 5. Finally, we draw our conclusions in Section 6.

\section{Background}




\vspace{-0.2cm}
\subsection{Adversarial Perturbation}

As mentioned, attacking a network means attempting to change the network decision on a particular target.
A target $t$ is defined as a region in the input
image where 
the generated adversarial perturbation is added to change the
network decision corresponding to this region.
For example, the target
$t$ for attacking an image classification network is the entire image.

The adversarial perturbation concept was first introduced for
attacking image classification networks in~\cite{intriguing,FGSM,delving,easilyfooled,deepfool,adversariaPhysical,universal}. 
Szegedy~\etal\cite{intriguing} showed that by adding
imperceptible perturbations to the input images, one could make the Convolutional Neural Network (CNN) predict the wrong class label
with high confidence. 
Goodfellow~\etal\cite{FGSM} explained that the vulnerability of
the neural networks to the adversarial perturbations is caused by
the linear nature of the neural networks. They proposed a fast
method to generate such adversarial perturbations, naming the method Fast
Gradient Sign Method (FGSM) defined by:
$\Mat{\xi} = \alpha \operatorname{sign}(\nabla_\Mat{X}\ell(f(\Mat{X}),y^{true}))$,
\noindent
where $\alpha$ was a hyper-parameter~\cite{FGSM}. The gradient
was computed with respect to the entire input image $\Mat{X} \in \mathbb{R}^{w \times h}$ by
back-propagation and the function $\operatorname{sign()}$ is
the $L_{\infty}$ norm.
Following this, Kurakin~\etal\cite{adversariaPhysical} proposed to
extend FGSM by iteratively generating the adversarial
perturbations. 
At each iteration, the values of the perturbations
were clipped to control perceptibility.
We denote it as I-FGSM in this work.
To reduce perceptibility, Moosavi-Dezfooli~\etal\cite{deepfool} proposed the method DeepFool, which iteratively adds
the minimal adversarial perturbations to the images by assuming
the classifier was linear at each iteration. 
The existence of universal perturbations for
image classification was shown in~\cite{universal}.

More recently, adversarial examples were extended into various 
applications such as semantic
segmentation~\cite{DAG,universaSegmentation,houdini,adversarialSSICLRwl}
and object detection~\cite{DAG}.
Metzen~\etal\cite{universaSegmentation} adapted an approach described in 
\cite{adversariaPhysical} into the semantic segmentation domain, where every
pixel was a target. 
They demonstrated that the gradients of the
loss for different target pixels might point to the opposite
directions. 

In object detection, the instances of interest are the detected
objects.
Thus, the targets are the detected region proposals
containing the object.
An approach for generating adversarial perturbations for
object detection is proposed in~\cite{DAG}.
They claimed that generating adversarial perturbations in object
detection was more difficult than in the semantic segmentation task.
In order to successfully attack a detected object, one needs to
ensure all the region proposals associated with the
object/instance are successfully attacked. 
For example, if only $K$ out of $R$ region proposals are successfully
attacked, the detector can still detect the object by using the
other high-confidence-score region proposals that are not successfully
attacked. 


We note that all of the above approaches use whole image perturbations which have
the same size as the input image.
This is because these perturbations are generated by calculating the gradient with respect to the entire image. 
Thus, a generated perturbation for one
target may disrupt the perturbations generated for other targets.
To contrast these methods with our work, we categorize these methods as \textbf{IM}age based \textbf{P}erturbation (IMP) methods.

\subsection{Loss function}


In general, the perturbations are generated by optimizing a specific objective function.
Let $\mathcal{L} = \sum_{i=1}^T \mathcal{L}_{t_{i}}$ be the loss
function to optimize. The objective function is defined
as follows:
\setlength{\belowdisplayskip}{2pt}
\setlength{\abovedisplayskip}{2pt}
\begin{equation}
\scalebox{0.9}{$\argmin_{\Mat{\xi}} \sum_{i=1}^T \mathcal{L}_{t_{i}}( \Mat{\xi} ) \textrm{ ,}$}
\end{equation}
\noindent
where $T$ is the number of targets; $\mathcal{L}_{t_i}$
is the loss function for each individual target ${t_i}$; and $\Mat{\xi} \in \mathbb{R}^{w \times h}$ is the
adversarial perturbation which will be added into the input image
$\Mat{X}$.

According to the goals of adversarial attacks, the attacks can
be categorized into \textit{non-targeted adversarial
attacks}~\cite{FGSM,universal,DAG} and \textit{targeted adversarial attacks}~\cite{adversariaPhysical,universaSegmentation}.
For non-targeted adversarial attacks, the goal is to reduce the
probability of truth class $y^{true}$ of the given target $t$
and to make the network predict any arbitrary class, whereas the
goal of targeted adversarial attacks is to ensure the network
predict the target class $y^{target}$ for the target $t$. The
objective function of the targeted attacks can be summarized into the following formula:
\setlength{\belowdisplayskip}{3pt}
\setlength{\abovedisplayskip}{3pt}
\begin{align}
\scalebox{0.9}{$\argmin_{\Mat{\xi}} \mathcal{L}_{t_i}=\ell(f(\Mat{X}+\Mat{\xi},
t_i),y^{target}) -\ell(f(\Mat{X}+\Mat{\xi},
t_i),y^{true}) \textrm{ ,}$}
\label{eq:adv_loss_detect}
\end{align}
\noindent
where, $\Mat{\xi}$ is the optimum adversarial perturbation;
$f$ is the network classification score matrix on the target region; and $\ell$ is the network loss function.


In general, the face detection problem is considered as a binary
classification problem, which aims at classifying a region as face
($+1$) or non-face ($-1$) (\ie $y^{target}=\{+1,-1\}$). However,
in order to detect faces in various scales, especially for tiny
faces, recent face detectors utilizing Single-Stage
networks~\cite{Tinyfaces,SSH,S3FD} divide the face detection problem
into multiple scale-specific binary classification problems, and
learn their loss functions jointly. The objective function to
attack such a network is defined as:
\setlength{\belowdisplayskip}{3pt}
\setlength{\abovedisplayskip}{3pt}
\begin{equation}
\scalebox{0.9}{$\argmin_{\Mat{\xi}} \quad \mathcal{L}_{t_i}=\sum_{j=1}^S
\ell_{s_{j}}(f_{s_{j}}(\Mat{X}+\Mat{\xi}, t_i),y^{target})  \textrm{ ,}$}
\label{eq:adv_loss_face}
\end{equation}
\noindent
where, $S$ is the number of scales; and $\ell_{s_j}$ is the
scale-specific detector loss function. Compared to
Eq.~\ref{eq:adv_loss_detect}, the above objective is more
challenging. This is because a single face can not only be
detected by multiple region proposals/targets, but also be
detected by multiple scale-specific detectors. Thus, one can only
successfully attack a face when the adversarial perturbation fools all the scale-specific detectors.
In other words, attacking the single-stage face detection network is more challenging than the work in object detection~\cite{DAG}.

Finally, as our main aim is to prevent faces being detected, then our objective function is formally defined as:
\setlength{\belowdisplayskip}{3pt}
\setlength{\abovedisplayskip}{3pt}
\begin{equation}
\scalebox{0.9}{$\mathcal{L}=\sum_{i=1}^T \mathcal{L}_{t_i}=\sum_{i=1}^T\sum_{j=1}^S \ell_{s_{j}}(f_{s_{j}}(\Mat{X}+\Mat{\xi},
t_i),-1)  \textrm{ .} $}
\label{eq:loss face detection}
\end{equation}
\noindent
In this work, we use the recent state-of-the-art
Single-Stage face detector, HR~\cite{Tinyfaces}, which
jointly learns 25 different scale-specific detectors, \ie $S=25$.
Note that since our aim is to fail detection, we set
the target label as $-1$.

\section{Instance Perturbations Interference}
\label{sec:IPI}

When performing an attack using the existing adversarial perturbation approaches~\cite{adversariaPhysical,universaSegmentation}, the Instance Perturbations Interference (IPI) problem appears when multiple faces exist in the input image.
In short, the IPI problem refers to the conditions where
successfully attacking one instance of interest can reduce the
chance of attacking the other instances of interest. For the face
detection task, the instance of interest is a face. 
If not addressed, the IPI problem will significantly reduce the overall attack success rate.

To show the existence of the IPI problem, we perform an experiment using synthetic images. 
In this experiment, we apply an adaptation of the existing perturbation methods 
generated by minimizing Eq.~\ref{eq:loss face detection}.

\subsection{Image based perturbation}
\label{sec:IP}

As mentioned, we categorize the previous methods as IMage based Perturbation (IMP) as they use whole image perturbation to perform the attack.
Here we adapt two of the existing methods, I-FSGM~\cite{adversariaPhysical} and DeepFool~\cite{deepfool}, by optimizing Eq.~\ref{eq:loss face detection}.
We denote them as IMP(I-FGSM) and IMP(DeepFool).
In both methods, the adversarial perturbation
is generated by using a gradient descent approach. 
At the ${(n+1)}$th
iteration, the gradient with respect to the input image $\Mat{X}$,
\scalebox{0.9}{$\nabla_\Mat{X}\mathcal{L}(f(\Mat{X}+\Mat{\xi}^{(n)}),-1)$}, is generated via
back-propagating the network with the loss function.

For the IMP(I-FSGM)~\cite{adversariaPhysical}, we iteratively
update the adversarial perturbation as follows:
\setlength{\belowdisplayskip}{3pt}
\setlength{\abovedisplayskip}{3pt}
\begin{align}
\scalebox{0.9}{$\Mat{\xi}^{(n+1)}=\operatorname{Clip_{\varepsilon}}\{\Mat{\xi}^{(n)}-\alpha\operatorname{sign}(\nabla_\Mat{X}\mathcal{L}(f(\Mat{X}+\Mat{\xi}^{(n)}),-1))\} $}
\textrm{ ,} \label{eq:iterative method wrt x}
\end{align}
\noindent
where the step rate $\alpha=1$; the epsilon $\varepsilon$ is the
maximum absolute magnitude to clip; \scalebox{0.9}{$\Mat{\xi}^{(0)}=\Mat{0}$}; and the loss function $\mathcal{L}$ is referred to the Eq.~\ref{eq:loss
face detection}. Note that in Eq.~\ref{eq:loss
face detection}, the loss function is a summation of the loss of
all targets. Thus, the aggregate gradient,
$\nabla_\Mat{X}\mathcal{L}$, can be rewritten as:
\setlength{\belowdisplayskip}{3pt}
\setlength{\abovedisplayskip}{3pt}
\begin{equation}
\scalebox{0.9}{$\nabla_\Mat{X}\mathcal{L}(f(\Mat{X}+\Mat{\xi}^{(n)}),-1)= \sum_{i=1}^T\sum_{j=1}^S
\nabla_\Mat{X}\ell_{s_{j}}(f_{s_{j}}(\Mat{X}+\Mat{\xi}^{(n)},t_i),-1)  \textrm{ .} $}
\label{eq:gradient is sum}
\end{equation}

As we assume $f$ is a deep neural network, then the
aggregate gradient \scalebox{0.9}{$\nabla_\Mat{X}\mathcal{L}$} can be obtained by
back-propagating all of the targets at once.
After obtaining the final adversarial perturbation $\Mat{\xi}$, the perturbed image, $\Mat{X}^{adv}$, is then generated by:
\scalebox{0.9}{$\Mat{X}^{adv}= \Mat{X}+\Mat{\xi}$}.

For the IMP(DeepFool), following~\cite{deepfool}, we configure the Eq.~\ref{eq:iterative method wrt x} into:
\setlength{\belowdisplayskip}{3pt}
\setlength{\abovedisplayskip}{3pt}
\begin{equation}
\scalebox{0.9}{$\Mat{\xi}^{(n+1)}=\operatorname{Clip_{\varepsilon}}\{\Mat{\xi}^{(n)}-\frac{\nabla_\Mat{X}\mathcal{L}(f(\Mat{X}+\Mat{\xi}^{(n)}))}{\norm{\nabla_\Mat{X}\mathcal{L}(f(\Mat{X}+\Mat{\xi}^{(n)}))}^2_2}\}
\textrm{ ,} $}
\label{eq:iterative method wrt x: DeepFool}
\end{equation}
where the loss function in Eq.~\ref{eq:loss face detection} is rewritten as \scalebox{0.9}{$\mathcal{L}=\sum_{i=1}^T\sum_{j=1}^S (f_{s_{j}}(\Mat{X}+\Mat{\xi},
t_i))$}.

Compare with the IMP(DeepFool), the IMP(I-FGSM) generates denser and more perceptible perturbations due to the $L_\infty$ norm.

\subsection{Existence of the IPI problem}

\begin{figure}[t]
\begin{center}
   \includegraphics[width=0.65\linewidth]{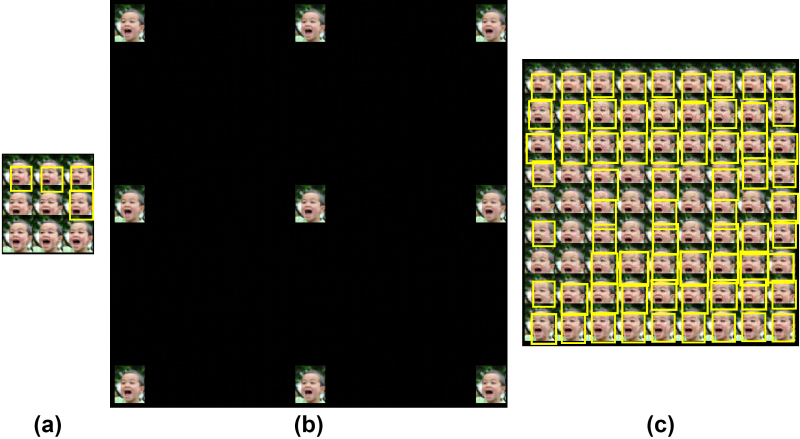}
\end{center}
   \caption{Examples of synthetic images after adding adversarial perturbations from the IMage based Perturbation (IMP). The detection results of the adversarial images are shown in rectangles. Note that, as face density increases, attack success rate decreases. The IMP attack is ineffective when there are many faces in an image, as in (a) and (c). When the distance among faces is increased, the attack becomes successful as in (b). }
\label{fig:example_syn}
\end{figure}

To show the existence of the IPI problem, we construct a set of synthetic images by controlling the number of 
faces and distances between them:
(1) an image containing only one face;
(2) an image containing multiple faces closely located in a grid and (3) using image (2) but increasing the distance between the faces. 
Examples are shown in Fig.~\ref{fig:example_syn}.
For this experiment, we use the recent state-of-the-art face
detector HR-ResNet101~\cite{Tinyfaces}.
The synthetic images are constructed by randomly selecting 50 faces from the WIDER FACE dataset~\cite{WiderFace}.
Experimental details are given in Section~\ref{sec:synthetic}.
We generate the adversarial perturbations using the IMP approaches: IMP(I-FGSM) and IMP(DeepFool).


\begin{table}[t]
\begin{center}
\subfloat[][IMP(I-FGSM)]{
\begin{tabular}{l|c|c c c|  c}
\hline
 N &1 & \multicolumn{3}{|c|}{16} 
 & 81 \\ 
 \hline
 Distance &40 &40 &80 & 160 &40 \\
 \hline
IMP(I-FGSM) &100 &34 &37.5 &38.9 &18.3
 \\
\hline
\end{tabular}}
\qquad
\subfloat[][IMP(DeepFool)]{
\begin{tabular}{l|c|c c c|  c}
\hline
 N &1 & \multicolumn{3}{|c|}{16} 
 & 81 \\ 
 \hline
 Distance &40 &40 &60 & 80 &40 \\
 \hline
IMP(DeepFool) &100 &67.5 &91.8 &99.7 &11.0
 \\
\hline
\end{tabular}}
\end{center}
\caption{The IMP attack success rate (in $\%$) on the synthetic images with respect to the number of faces and distances among faces. $N$ is the number of faces. The IMP can achieve $100\%$ attack success rate when there is one face per image. The attack success rate drops significantly when the number of faces is increased. With the same number of faces, the attack success rate can be increased as the distance among faces increases.  }
\label{tab:IPI syn}
\end{table}


The attack success rate is calculated as follows: 
$\frac{\operatorname{\#Face\ removed} }{ \operatorname{\#Detected\
face}}$. 
Table~\ref{tab:IPI syn} reports the results. For the first synthetic case
where an image only contains one face, both IMP(I-FGSM) and IMP(DeepFool) are
able to attack the face detector with a $100\%$ attack success rate. 
The IMP method is only
partially successful on the second case where the number of faces is increased to $16$. 
The attack success rates decrease significantly to only $18.3\%$ and $11.0\%$ when $N=81$.  
The IMP method attack success rates significantly increase 
when the distances between faces are increased significantly, especially for the IMP(DeepFool).
It is because the IMP(DeepFool) generates sparser perturbations than the IMP(I-FGSM).

These results suggest the following: (1) IMP is
effective when only a single face exists; (2) IMP is
ineffective when multiple faces exist close to each other and (3)
the distance between faces significantly affects the attack
performance. There are two questions that arise from these
results: (1) why is the attack affected by the number of faces?
and (2) why does the distance between faces affect the attack success
rate? We address these two questions in the next section.







\section{Proposed method}

We first elaborate on the relationship between the Effective Receptive Field and the IPI problem. Then, the proposed Localized Instance Perturbation (LIP) method is outlined.

\subsection{Effective Receptive Field (ERF)}
\label{sec:ERF}


The receptive field of a neuron in a neural network is the
set of pixels in the input image that impact the neuron decision~\cite{ERF}.
In CNNs, it has been shown in~\cite{ERF} that
the distribution of impact within the Theoretical Receptive Field (TRF) of a neuron follows a 2D Gaussian distribution in many cases.
This means most pixels that have significant impact to
the neuron decision are
concentrated near the neuron.
In addition, the impact decays quickly away from the center of the TRF.
In~\cite{ERF}, the area where pixels still have significant impact to the neuron decision is defined as the Effective Receptive Field (ERF).
The ERF only takes up a fraction of the TRF and pixels within the ERF will generate non-negligible impacts on the final outputs.
We argue that understanding ERF and TRF is important for addressing the IPI problem.
This is because the adversarial perturbation is aimed at changing a network decision at one or more neurons.
This means, all pixels in the input image that impact the decision must be considered when determining the perturbation.


In this paper, we denote the Distribution of Impacts in the TRF as DI-TRF for simplicity. The DI-TRF is measured by calculating the partial derivative of the central pixel on the output layer via back-propagation.
Following the notations in our paper, let us denote the central pixel as $t_c$, then the partial derivative of the central pixel is $\frac{\partial f(\Mat{X},t_c)}{\partial \Mat{X}}$, which is the DI-TRF.
According to the chain rule, we have the gradient of the target $t_c$~\cite{ERF} as: $\nabla_\Mat{X}\mathcal{L}(f(\Mat{X},t_c),y^{target})=\frac{\partial \mathcal{L}(f(\Mat{X},t_c),y^{target})}{\partial f(\Mat{X},t_c)}\frac{\partial f(\Mat{X},t_c)}{\partial \Mat{X}}$, 
where the $\frac{\partial \mathcal{L}(f(\Mat{X},t_c),y^{target})}{\partial f(\Mat{X},t_c)}$ is set to 1.

Comparing the gradient of a target pixel for the adversarial perturbations in Eq.~\ref{eq:gradient is sum}, the only difference with the DI-TRF is in the partial derivative of the loss function $\frac{\partial \mathcal{L}(f(\Mat{X},t_c),y^{target})}{\partial f(\Mat{X},t_c)}$, which is a scalar for one target pixel.
In our work, the scalar, $\frac{\partial \mathcal{L}(f(\Mat{X},t_c),y^{target})}{\partial f(\Mat{X},t_c)}$, measures the loss between the prediction label and the target label.
The logistic loss is used for the binary classification of each scale-specific detector, (\ie the $\ell_{s_j}(f_{s_{j}}(\Mat{X},t_c),y^{target})$ in Eq.~\ref{eq:loss face detection}).
Therefore, our adversarial perturbation for one target can be considered as a scaled distribution of the DI-TRF.
Since DI-TRF follows a 2D Gaussian distribution~\cite{ERF}, then the adversarial perturbation to change a single neuron decision is also a 2D Gaussian.


We explain the IPI problem as follows.
Since an adversarial perturbation to attack a single neuron follows a 2D Gaussian, then the perturbation is mainly spread over the ERF and will have a non-zero tail outside the ERF.
From the experiment, we observed that the perturbations generated to attack multiple faces in the image may interfere with other.
More specifically, when these perturbations overlap
with the neighboring face ERF, they may be sufficient enough to
disrupt the adversarial perturbation generated
to attack this neighboring face. 
In addition, prior work~\cite{mitigatingRandomization} shows that adversarial attacks might fail when the specific structure is destroyed.
In other words, when multiple attacks are applied simultaneously, these 
attacks may corrupt each other, leading to a lower attack rate. 
We name the part of a perturbation interfering with the other perturbations for other faces as the interfering perturbation.

This also explains why the IPI is affected by the distance between faces.
The closer the faces, the more chance the interfering perturbations with a larger magnitude overlap with the neighboring face ERF.
When distances between faces increase, the magnitude of the interfering perturbations that overlap with the neighboring ERFs may not be strong enough to disrupt attacks for neighboring faces.

\subsection{Localized Instance Perturbation (LIP)}
\label{sec:LIP}

To address the IPI problem, we argue that the generated adversarial perturbations of one instance should be exclusively confined within the instance ERF.
As such, we call our method as the Localized Instance Perturbation (LIP).

The LIP comprises two main components: (1) methods to eliminate any possible interfering perturbation and (2) methods to generate the perturbation.
\subsubsection{4.2.1 Eliminating the interfering perturbation}

To eliminate the interference between perturbations, we attempt to constrain the 
generated perturbation for each instance individually inside the ERF. 
Let us consider that an image $\Mat{X}$, with $w \times h$ pixels, contains $N$ instances $\{ \Vec{m}_i\}_{i=1}^N$.
Each instance $\Vec{m}_i$ has its corresponding ERF, $\Vec{e}_i$, and we have $\{ \Vec{e}_i\}_{i=1}^N$.
For each instance, there are a set of corresponding targets represented as object proposals, $\{ p_j\}_{j=1}^P$.
We denote the final perturbation for the $i$th instance as $\Mat{R}_{m_i}$ and the final combination of the perturbation of all the instances as $\Mat{R}$.
Similar to the IMP method, once the final perturbation, $\Mat{R}$, has been computed, then we add the perturbation into the image \scalebox{0.9}{$\Mat{X}^{adv} = \Mat{X} + \Mat{R}$}.




\noindent
\textbf{Perturbation cropping -- }
This step is to limit the perturbations inside the instance ERF. 
This is done by cropping the perturbation according to the corresponding instance ERF.
Let us define a binary matrix $\Mat{C}_{\Vec{e}_i} \in \{0,1\}^{w \times h}$ as the cropping matrix for the ERF, $\Vec{e}_i$.
The matrix $\Mat{C}$ is defined as follows:
\setlength{\belowdisplayskip}{2pt}
\setlength{\abovedisplayskip}{3pt}
\begin{equation}
\scalebox{0.9}{$\Mat{C}_{\Vec{e}_i}(w,h) =
\begin{cases}
1, & (w,h)\in\Vec{e}_i\\
0, &\text{otherwise}
\end{cases}
\textrm{ ,} $}
\end{equation}
where $(w,h)$ is a pixel location.
The cropping operation is computed by a element-wise dot product of the mask $\Mat{C}_{\Vec{e}_i}$ and the gradient w.r.t. the input images $\Mat{X}$, is defined as:
\setlength{\belowdisplayskip}{3pt}
\setlength{\abovedisplayskip}{3pt}
\begin{equation}
\scalebox{0.9}{$\Mat{R}_{m_i}=\operatorname{C}_{\Vec{e}_i} \cdot \nabla_\Mat{X}\mathcal{L}_{m_i} \textrm{ ,} $}
\label{eq:instance_perturbation}
\end{equation}
\noindent
where $\mathcal{L}_{m_i}$ is the loss function of the $i$-th instance.
$\mathcal{L}_{m_i}$ will be described in the next sub-section.


\noindent
\textbf{Individual instance perturbation -- }
It is possible to compute the perturbation of multiple instances simultaneously. 
However, the interfering perturbation can still exist and may impact the attack.
To that end, we propose to separately compute the perturbation for each instance, $\nabla_\Mat{X}\mathcal{L}_{m_i}$ before cropping.

After the cropping step is applied to each instance perturbation, the final perturbation of all instances is combined via:
\setlength{\belowdisplayskip}{3pt}
\setlength{\abovedisplayskip}{3pt}
\begin{equation}
\scalebox{0.9}{$\Mat{R}=\sum_{i=1}^N \Mat{C}_{\Vec{e}_i}\cdot\nabla_\Mat{X}\mathcal{L}_{m_i} \textrm{ .} $}
\label{eq:sum R LIP}
\end{equation}
We then normalize the final perturbation, $\Mat{R}$, via: \scalebox{0.9}{$\Mat{R}=\alpha \operatorname{sign}(\Mat{R})$}.



\subsubsection{4.2.2 Perturbation generation}

Given a set of region proposals corresponding to an instance, there are at least two methods of generating the instance perturbation $\Mat{R}_{m_i}$: 
(1) All proposal based generation and (2) Highest Confidence proposal based generation.

\noindent
\textbf{All proposal based generation -- } In the first method, we utilize all the region proposals to generate the perturbation $\Mat{R}_{m_i}$.
Thus, the $\mathcal{L}_{m_i}$ in Eq.~\ref{eq:instance_perturbation} can be defined as a summation of the loss function of all the 
region proposals $\mathcal{L}_{p_j}$ belong to the instance:
\setlength{\belowdisplayskip}{3pt}
\setlength{\abovedisplayskip}{3pt}
\begin{equation}
\scalebox{0.9}{$\mathcal{L}_{m_i}=\sum_{j=1}^P \mathcal{L}_{p_j} \textrm{ .} $}
\label{eq:all pixels}
\end{equation}

\noindent
\textbf{Highest confidence proposal based generation -- }  In online hard example mining~\cite{OHEM}, Shrivastava~\etal showed the efficiency of using the hard examples to generate the gradients for updating the networks.
The hard examples are the high-loss object proposals chosen by the non-maximum suppression.
Non-Maximum Suppression (NMS) is similar to max-pooling, which selects the object proposal with the highest score (\ie selecting the proposal with the highest loss).

Inspired by this, instead of attacking all of the object proposals corresponding to a single instance, we can use NMS to select the one with highest loss to compute the back-propagation.
Then $\mathcal{L}_{m_i}$ can be rewritten as:
\setlength{\belowdisplayskip}{3pt}
\setlength{\abovedisplayskip}{3pt}
\begin{equation}
\scalebox{0.9}{$\mathcal{L}_{m_i}=\max (\mathcal{L}_{p_j}) \textrm{ .} $}
\label{eq:one pixel}
\end{equation}



\section{Experiments}
In this section, we first describe the implementation details and then evaluate our proposed adversarial attacks on the state-of-the-art face detection datasets.

\subsection{Implementation Details}

\begin{figure}
\begin{center}
   \includegraphics[width=0.6\linewidth]{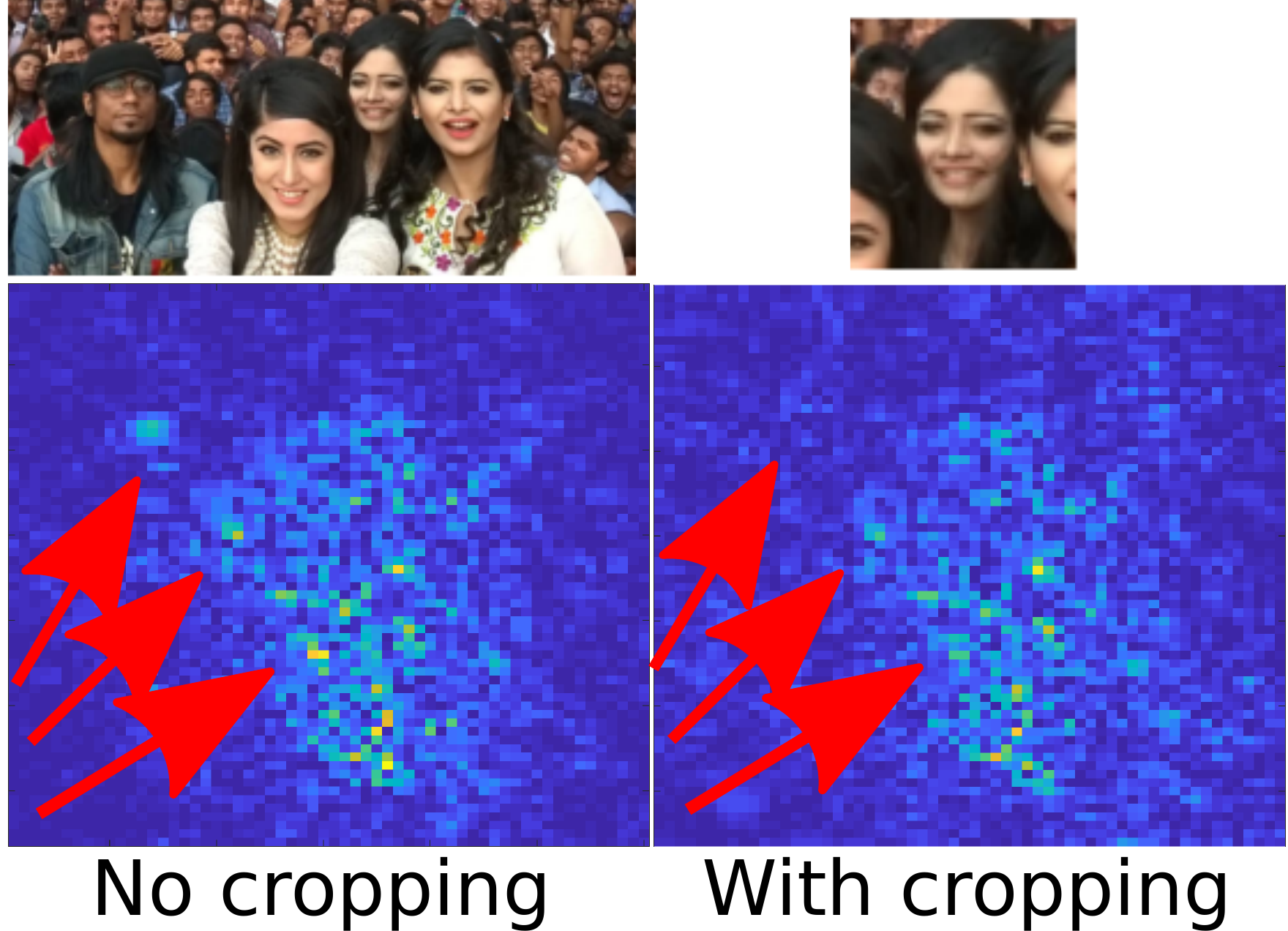}
\end{center}
   \caption{Example of non-normalized generated perturbations in absolute value. Left: without cropping the image; Right: with cropping. The red arrows point at differences between the two perturbations. Note that the context information is lost when cropped image is used (first arrow from the left). Please view in color.}
\label{fig:fig_context}
\end{figure}

For this study, we utilize recent state-of-the-art face detector, HR~\cite{Tinyfaces}.
In particular, HR-ResNet101 is used.
Image pyramids are utilized in HR, \ie downsampling/interpolating the input image into multiple sizes.
Therefore, for every image in the pyramid, we generate corresponding adversarial examples.
The detection results of the image pyramid are combined together with Non-Maximum Suppression (NMS).
The chosen thresholds of NMS and classification are $0.1$ and $0.5$ respectively.


In order to avoid the gradient explosion when generating the perturbations, we found that by zero-padding the small input images can reduce the magnitude of the gradient.
In this work, we zero pad the small images to $1000\times1000$ pixels.
In addition, as the input images of the detection networks can have arbitrary sizes, we do not follow existing methods~\cite{universal,universaSegmentation} that
resize the input images into a canonical size.

Note that we cannot simply crop the input image to generate a successful adversarial perturbation.
This is because the perturbation may be incomplete as it does not include the context information obtained from neighboring instances.
An example of two non-normalized perturbations in absolute value generated with and without context is shown in Fig.~\ref{fig:fig_context}~\footnote{These perturbation are not normalized for the sake of illustration.}.


For determining the perturbation cropping size, we follow the work of Luo~\etal\cite{ERF} which computes the gradient of the central proposal of an instance on the output feature map to obtain the distribution of the ERF.
We average the gradients over multiple instances and determine the crop size with the definition that the ERF takes up $90\%$ of the energy of the TRF~\cite{ERF}.
The perturbation crop size is set to $80\times80$ pixels for small faces and $140\times140$ pixels for large faces.
The maximum noise value $\varepsilon$ is $20$ and the maximum number of iterations $N_0$ is $40$.
The $\alpha$ is set to 1 in this work.

\medskip
\noindent
\textbf{Perturbation Generation Methods -- } In our work, we compared our proposed Localized Instance Perturbation (LIP) approach with the IMage Perturbation (IMP) and Localized Perturbation (LP).
The details of the perturbation generation methods evaluated are listed as follow:

\noindent
(1) \textbf{Localized Instance Perturbation using All proposal generation (LIP-A) -- } The proposed LIP-A is a variant of our proposed LIP method in Section~\ref{sec:LIP}.
As mentioned, the loss function of one instance is the summation of all proposals (refer to Eq.~\ref{eq:all pixels}).

\noindent
(2) \textbf{Localized Instance Perturbation using Highest confidence proposal generation (LIP-H) -- } The LIP-H is another variant of our proposed LIP with the loss function of Eq.~\ref{eq:one pixel}. 
The loss function of one instance consists of only one loss of the highest confidence proposal.

\noindent
(3) \textbf{IMage Perturbation (IMP) -- } The IMP method refers to the generation method in Section~\ref{sec:IP} which applies the perturbation without cropping it.
This perturbation generation method follows the previous work~\cite{universaSegmentation}.

\noindent
(4) \textbf{Localized Perturbation (LP) -- } The LP is the localized perturbation which also crops the image perturbation. 
The main difference to the proposed LIP is that it computes the gradients of all the instances simultaneously before the cropping.
In contrast to Eq.~\ref{eq:sum R LIP}, the final perturbation is obtained by:
\setlength{\belowdisplayskip}{2pt}
\setlength{\abovedisplayskip}{2pt}
\begin{equation}
\scalebox{1}{$\Mat{R}= \bigcup_{i=1}^{N}\Mat{C}_{\Vec{e}_i}\cdot \sum_{i=1}^N\nabla_\Mat{X}\mathcal{L}_{m_i} \textrm{ .}$}
\label{eq:sum R LP}
\end{equation}
\noindent
where $\bigcup_{i=1}^{N}\Mat{C}_{\Vec{e}_i}$ is the union of all binary matrices.
The advantage of this method is that current deep learning toolboxes can calculate the summation of the gradients of all instances, (\ie $ \sum_{i=1}^N\nabla_\Mat{X}\mathcal{L}_{m_i}$), simultaneously by back-propagating the network only once.

\medskip
\noindent
\textbf{Benchmark Datasets -- } We evaluate our proposed adversarial perturbations on two recent popular face detection benchmark datasets:
(1) \textbf{FDDB dataset~\cite{FDDB}:} 
The FDDB dataset is designed to benchmark face detectors in unconstrained environments.
The dataset includes images of faces with a wide range of difficulties such as occlusions, difficult poses, low resolution and out-of-focus faces.
It contains 2,845 images with a total of 5,171 faces labelled; and (2) \textbf{WIDER FACE dataset~\cite{WiderFace}:} 
The WIDER FACE dataset is currently the most challenging face detection benchmark dataset.
It comprises 32,203 images and 393,703 annotated faces based on 61 events collected from the Internet.
The images of some events, \eg parade, contain a large number of faces.
According to the difficulties of the occlusions, poses and scales, the faces are grouped into three sets: 'Easy', 'Medium' and 'Hard'.
As for the experiments, we randomly choose $1,000$ images from the validation set.

\medskip
\noindent
\textbf{Evaluation Metrics -- }
The metrics for evaluating the adversarial attacks against face detection are attack success rate and detection rate defined as follows:
(1) \textbf{Attack Success Rate:}  The attack success rate is the ratio between the number of faces that are successfully attacked and the number of detected faces before the attacks; and (2) \textbf{Detection Rate:} The detection rate is the ratio between the number of detected faces and the number of faces in the images.

\subsection{Evaluation on synthetic data}
\label{sec:synthetic}
As discussed in Section~\ref{sec:IPI}, due to the IPI problem, the IMP does not perform well on the cases where (1) the number of faces per image is large; and (2) the faces are close to each other.
Here, we contrast IMP with LP and LIP.


We randomly selected $50$ faces from the WIDER FACE dataset~\cite{WiderFace}. 
These faces were first resized into a canonical size of $30 \times 30$ pixels.
Each face was then duplicated and inserted into a blank image in a rectangular grid manner (\eg $3 \times 3 = 9$).
The number of duplicates and the distance between the duplicates were controlled during the experiment.
In total there were $50$ images and the attack success rate was then averaged across $50$ images.
Some examples of the synthetic images are shown in Fig.~\ref{fig:example_syn}. 

\setlength{\abovecaptionskip}{0pt}
\begin{figure}[t]
\begin{center}
   \includegraphics[width=1\linewidth]{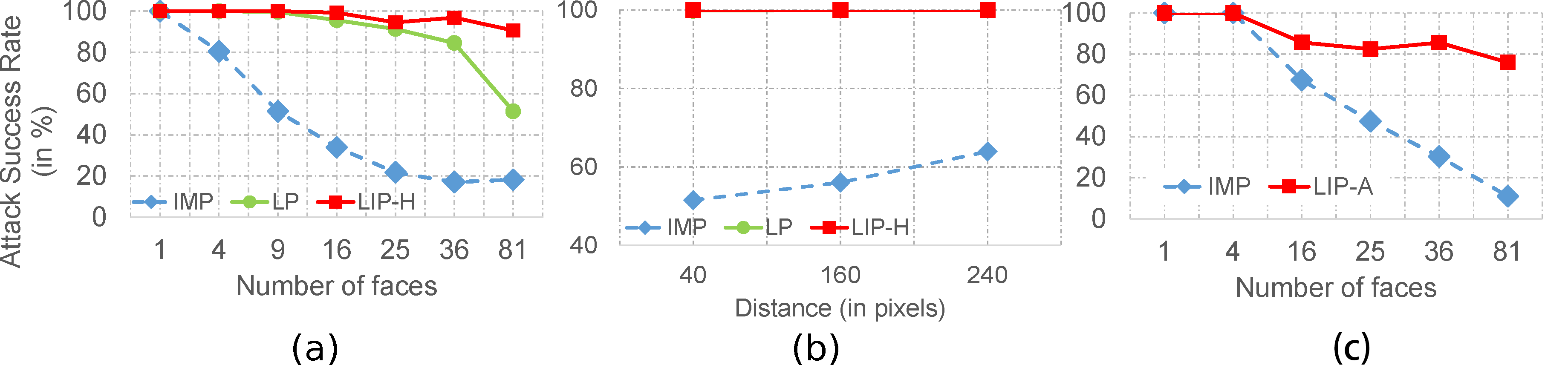}
\end{center}
   \caption{The attack success rate of the I-FGSM with respect to: (a) the number of faces. The distance was fixed to 40 pixels; and (b) the distance between faces. Nine face duplicates were used. (c) The attack success rate of the DeepFool. }
\label{fig:syn_Distance_Nface}
\end{figure}


\noindent
\textbf{The effect of the number of faces -- } we progressively increased the number of duplication for each synthetic image 
from $1 \times 1$ to $9 \times 9 = 81$ duplicates. We fixed the distance between duplicates to 40 pixels.
The quantitative results are shown in Fig.~\ref{fig:syn_Distance_Nface}.
From this figure, we can see that for the perturbation generation method I-FGSM, the IMP attack success rate significantly drops from $100\%$ to $20\%$ as the number of faces is increased.
On the contrary, both LP and LIP-H can achieve significantly higher attack success rate than IMP.
This is because both LP and LIP-H only use the generated perturbation within the corresponding instance ERF by cropping it before applying.
Note that, when the number of faces is more than $36$, the LP attack success rate drops from $85\%$ ($N=36$) to $51\%$ ($N=81$), 
whereas the LIP-H can still achieve more than $90\%$ success rate.
Since LP processes all the instances simultaneously, 
the accumulation of the interfering perturbations within each instance ERF will become more significant when the number of faces is increased.
Similarly, for the generation method DeepFool, the LIP has demonstrated its effectiveness on addressing the IPI problem when multiple faces exist.
These also suggest the existence of the IPI problem.

\noindent
\textbf{The effect of distance between faces -- } In this experiment the number of faces duplication was fixed to $9$. We modified the distance between face duplicates to 40, 160 and 240 pixels.
It can be seen from Fig.~\ref{fig:syn_Distance_Nface}b that the attack success rate for IMP increases as the distance between faces is increased.
The performance of both LP and LIP-H are not affected.
Similar performance is achieved on the DeepFool. More details are shown in the supplementary materials due to the space limitation.






\subsection{Evaluation on face detection datasets}

We contrasted LIP-A and LIP-H with IMP and LP based on two existing methods: I-FSGM~\cite{adversariaPhysical} and DeepFool~\cite{deepfool}.
The experiments were run on the entire FDDB~\cite{FDDB} and 1,000 randomly selected images in the validation set of WIDER FACE~\cite{WiderFace}.

\begin{table}
\begin{center}
\begin{tabular}{l|C{0.9cm}|C{0.9cm}|C{0.9cm}|C{0.9cm}|C{0.9cm}}
\hline
\multirow{2}{*}{Perturbations}  &\multirow{2}{*}{none} &\multicolumn{4}{c}{I-FGSM}\\
\cline{3-6}
 & &IMP &LP &LIP-A &LIP-H \\
\hline
Detection Rate &95.7 &44.1  &4.8 &5.1  &5.9 \\
\hline
Attack Success Rate  & -- &53.9&94.9  &94.6 &93.8\\
\hline
\end{tabular}
\end{center}
\caption{The attack success rates and detection rates (in \%) on FDDB~\cite{FDDB}.}
\label{tab:fddb}
\end{table}

\begin{table}
\begin{center}
\begin{tabular}{l|c|C{0.9cm}|C{0.9cm}|C{0.9cm}|C{0.9cm}|C{0.9cm}|C{0.9cm}|C{0.9cm}}
\hline
\multirow{2}{*}{Perturbations} & \multirow{2}{*}{} &\multirow{2}{*}{none} &\multicolumn{4}{c}{I-FGSM}   &\multicolumn{2}{|c}{DeepFool} \\
\cline{4-9}
 & & &IMP &LP &LIP-A &LIP-H &IMP  &LIP-A\\
\cline{1-9}
\multirow{3}{*}{Detection Rate} &easy &92.4 &46.2  &30.1 &28.2  &\textbf{26.5} &50.6  &43.2\\
\cline{2-9}
&medium &90.7 &50.7  &34.7 &32.2  &\textbf{31.1} &54.4  &40.0\\
\cline{2-9}
&hard &77.3 &45.9  &29.3 &\textbf{23.6}  &26.6 &46.5  &25.8\\
\hline
\multirow{3}{*}{Attack Success Rate} &easy &-- &50.0 &67.4 &69.5  &\textbf{71.3} &45.3  &53.2\\
\cline{2-9}
&medium &--&44.1  &61.7 & 64.5 &\textbf{65.7} &40.0  &56.4\\
\cline{2-9}
&hard &-- &40.6  &62.1 &\textbf{69.5} &65.6 &39.6  &66.6\\
\hline
\end{tabular}
\end{center}
\caption{The attack success rate and detection rate (in \%) on WIDER FACE validation set~\cite{WiderFace}.}
\label{tab:wider}
\end{table}

The results based on the I-FGSM, are reported in Table~\ref{tab:fddb} and Table~\ref{tab:wider} respectively.
On the FDDB dataset (in Table~\ref{tab:fddb}), the face detector, HR~\cite{Tinyfaces}, achieves $95.7\%$ detection rate.
The LP, LIP-A and LIP-H can significantly reduce the detection rate to around $5\%$ with the attack success rate of 
$94.9\%$, $94.6\%$ and $93.8\%$ respectively.
On the other hand, the IMP can only achieve $53.9\%$ attack success rate (\ie significantly lower than the LP, LIP-A, LIP-H performance).
This signifies the importance of the perturbation cropping to eliminate the interfering perturbations.
Due to the IPI problem, the interfering perturbations from the other instances will affect the adversarial attacks of the target instance. 
This results in the low attack success rate of the IMP.
This is because to generate the perturbations, the IMP simply sums up the all perturbations including the 
interfering perturbations.
We note that the performance of LP, LIP-A and LIP-H are on par in the FDDB dataset. 
This could be due to the low number of faces per image for this dataset.

\begin{figure}
\begin{center}
   \includegraphics[width=0.6\linewidth]{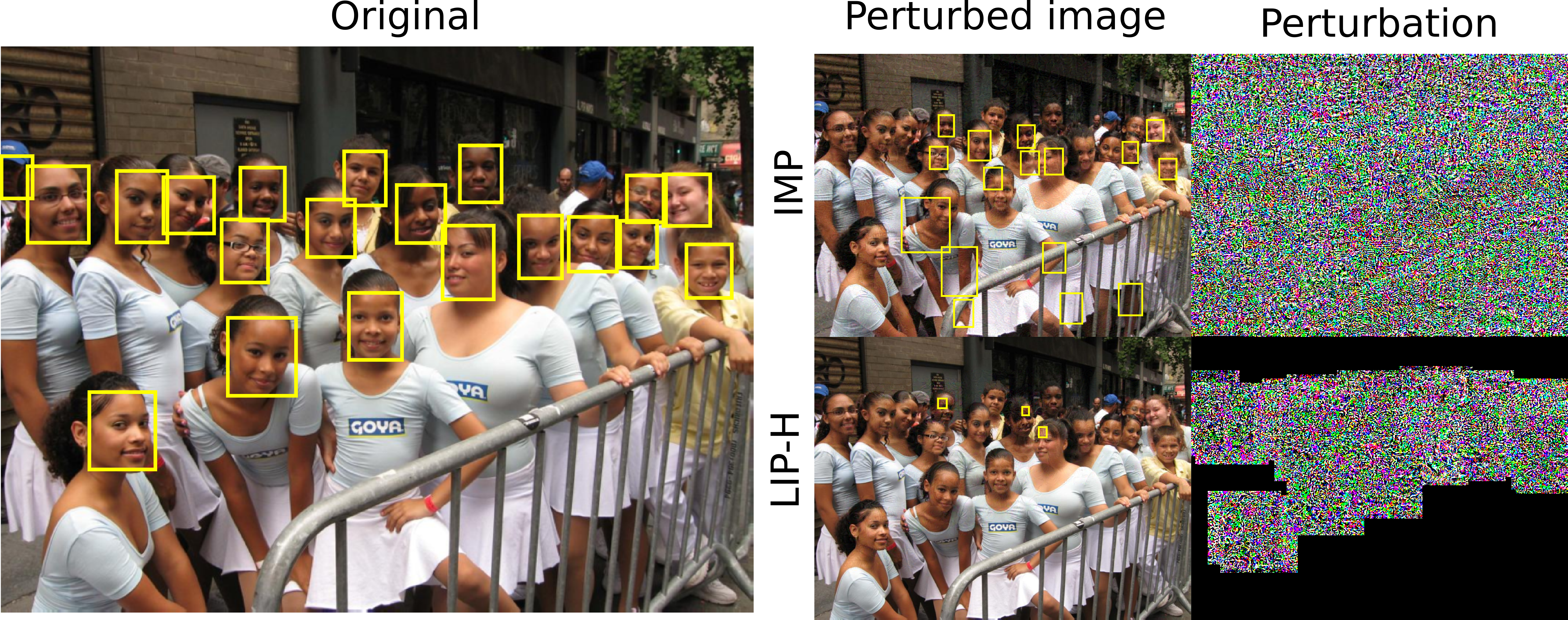}
\end{center}
   \caption{Examples of the adversarial attacks on face detection network, where the perturbation generation is based on the I-FGSM. The LIP-H is successfully attack all the faces, whereas some faces are still detected when IMP is used.}
\label{fig:perceptibility}
\end{figure}

However, when the number of faces per image increases significantly, LIP shows its advantages.
Examples can be seen in Fig.~\ref{fig:perceptibility}.
This can be observed in the WIDER FACE dataset where LIP-A and LIP-H outperform LP by 4 percentage points.
the LIP-H can achieve attack success rates of $(69.8\%, 63.7\%, 61.4\%)$ on the (easy, medium, hard) sets, while 
the LP can only obtain attack success rate $(65.7\%, 59.5\%, 57.4\%)$.
As the LP processes all the instances together, the interfering perturbations are accumulated within the ERF before the cropping step.
Note that the interfering perturbations may have low magnitude, however, when they are accumulated due to the number of neighboring instances then 
disruption could be significant.
These results also suggest that we do not necessarily need to attack all the region proposals as the performance of LIP-H is on par with LIP-A.

Similarly, for the DeepFool based methods, the LIP has demonstrated its effectiveness on addressing the IPI problem. Full experimental results are shown in the supplementary materials.

\section{Conclusions}
In this paper we presented an adversarial perturbation method to
fool a recent state-of-the-art face detector utilizing the
single-stage network. We described and addressed the Instance
Perturbation Interference (IPI) problem which was the root cause
for the failure of the existing adversarial perturbation
generation methods to attack multiple faces simultaneously. 
We found that it was sufficient to only use the generated perturbations 
within an instance/face
Effective Receptive Field (ERF) to perform an effective attack.
In addition, it was
important to exclude perturbations outside the ERF to avoid
disrupting other instance perturbations. 
We thus proposed the
Localized Instance Perturbation (LIP) approach that only confined
the perturbation within the ERF.
Experiments showed that the
proposed LIP successfully generated perturbations for multiple
faces simultaneously to fool the face detection network and
significantly outperformed existing adversarial generation
methods.
As currently the perturbations are generated specifically to each
face, we plan to develop a universal perturbation generation
method which can attack many faces with a general perturbation. 


\subsection{Acknowledgement}

This work has been funded by the Australian Research Council (ARC) Linkage Projects Grant LP160101797. Arnold
Wiliem is funded by the Advance Queensland Early-Career Research Fellowship.

\clearpage

\bibliographystyle{splncs}
\bibliography{reference}
\end{document}